\documentclass[11pt,a4paper]{article}
\usepackage[hyperref]{acl2018}
\usepackage{times}
\usepackage{latexsym}
\usepackage{amsmath}
\usepackage{multirow}
\usepackage{url}
\usepackage{enumitem}
\usepackage{todonotes}
\usepackage{booktabs}
\usepackage{siunitx}
\usepackage{tabularx,nonfloat}
\usepackage{color, colortbl}

\newcolumntype{L}{>{\raggedright\arraybackslash}X}

\definecolor{Restaurant}{rgb}{0.906,0.949,0.902}
\definecolor{Hotel}{rgb}{0.667,0.71,0.761}

\aclfinalcopy 
\def\aclpaperid{96} 

\def\experimentsgap{0pt}

\title{Addressing Objects and Their Relations: \\The Conversational Entity Dialogue Model}

\author{Stefan Ultes, Pawe\l\ Budzianowski, I\~nigo Casanueva, Lina Rojas-Barahona, \\\textbf{Bo-Hsiang Tseng, Yen-Chen Wu, Steve Young and Milica~Ga{\v s}i{\' c}} \\
  Cambridge University Engineering Department \\
  Cambridge, United Kingdom\\
   {\tt \{su259,pfb30,ic340,lmr46,bht26,ycw30,sjy11,mg436\}@cam.ac.uk} 
  \\}

\date{}

\begin{document}
\maketitle
\begin{abstract}
Statistical spoken dialogue systems usually rely on a single- or multi-domain dialogue model that is restricted in its capabilities of modelling complex dialogue structures, e.g., relations. In this work, we propose a novel dialogue model that is centred around entities and is able to model relations as well as multiple entities of the same type. We demonstrate in a prototype implementation benefits of relation modelling on the dialogue level and show that a trained policy using these relations outperforms the multi-domain baseline. Furthermore, we show that by modelling the relations on the dialogue level, the system is capable of processing relations present in the user input and even learns to address them in the system response.
\end{abstract}

\section{Introduction}
\label{sec:intro}
Data-driven statistical spoken dialogue systems (SDS)~\cite{lemon2012,young2013} are a promising approach for realizing spoken dialogue interaction between humans and machines. 
Up until now, these systems have successfully been applied to single- or multi-domain task-oriented dialogues~\cite{su-EtAl:2017:SIGDIAL,casanueva2017benchmarking,
lison2011multi,wang2014policy,papangelis2017single,
gasic2017,budzianowski2017,
peng2017composite} where each dialogue is modelled as multiple independent single-domain sub-dialogues. However, this multi-domain dialogue model (MDDM) does not offer an intuitive way of representing multiple objects of the same type (e.g., multiple restaurants) or dynamic relations between these objects. To the best of our knowledge, neither problem has yet been addressed in statistical SDS research.

The goal of this paper is to propose a new dialogue model---the conversational entity dialogue model (CEDM)---which offers an intuitive way of modelling dialogues and complex dialogue structures inside the dialogue system. Inspired by \citet{grosz1978}, the CEDM is centred around objects and relations instead of domains thus offering a fundamental change in how we think about statistical dialogue modelling. 
The CEDM allows
\begin{itemize}
\item to model dynamic relations directly, independently and persistently so that the relations may be addressed by the user \emph{and} the system,
\item the system to talk about multiple objects of the same type, e.g., multiple restaurants,
\end{itemize}
while still allowing feasible policy learning.

The remainder of the paper is organized as follows: after presenting a brief motivation and related work in Section~\ref{sec:related_work},
Section~\ref{sec:ssds} presents background information on statistical SDSs. Section~\ref{sec:cedm} contains the main contribution and describes the conversational entity dialogue model in detail. Looking at one aspect of the CEDM, the modelling of relations, Section~\ref{sec:evaluation} describes a prototype implementation and shows the benefits of the CEDM in experiments with a simulated user. Section~\ref{sec:conclusion} concludes the paper with a  list of open questions which need to be addressed in future work.

\begin{figure*}[t]
\centering
\includegraphics[width=0.9\linewidth]{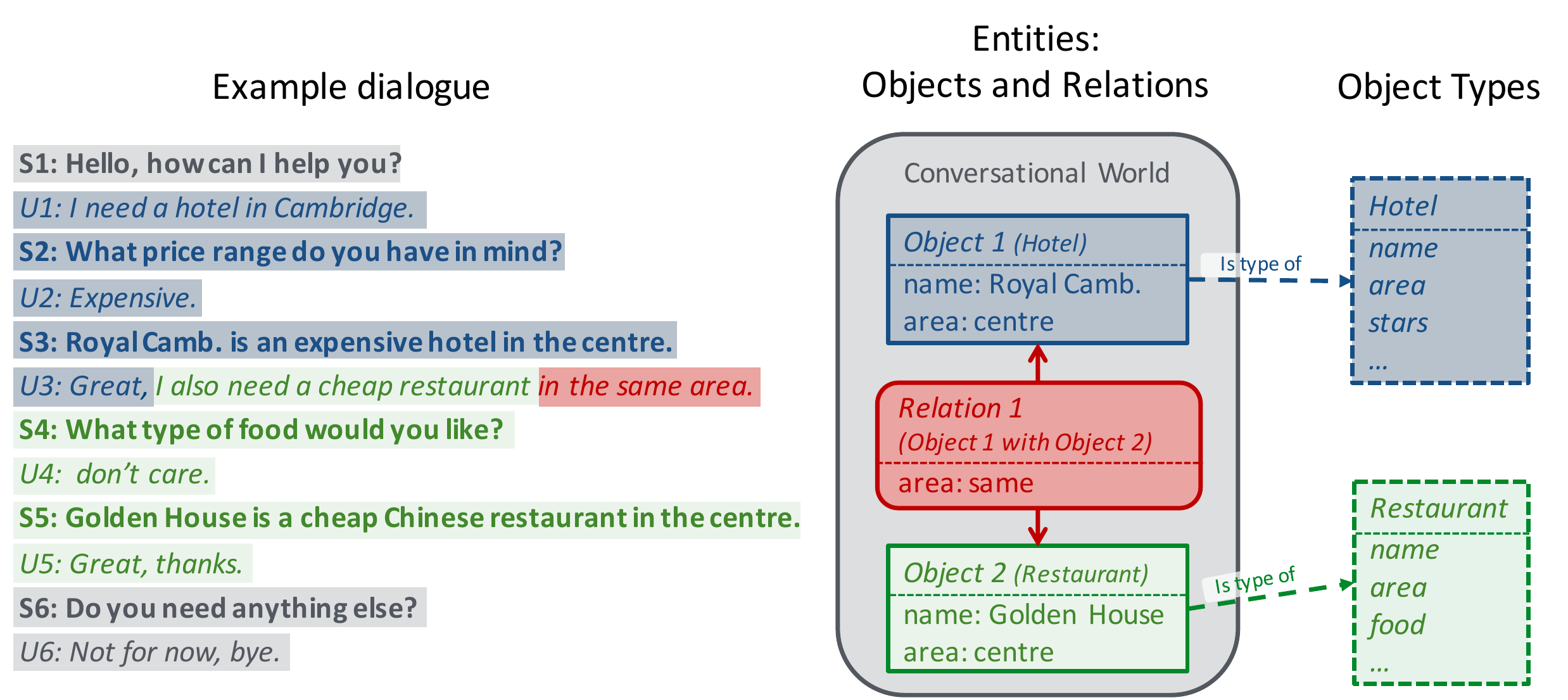}
\caption{A dialogue between the system (S) and a user (U) about a restaurant and a hotel \textit{in the same area} along with the mapping of fractions of the dialogue to the respective objects (of predefined types) and the relation. All objects and relations reside inside a conversational world.}
\label{fig:dialogue}
\end{figure*}

\section{Motivation and Related Work}
\label{sec:related_work}
To introduce the terminology that will be used in this work and to illustrate the necessity of adequate modelling of relations, Figure~\ref{fig:dialogue} shows an example dialogue about hotels and restaurants in Cambridge with the relation \textit{in the same area}. Instead of talking about a sequence of domains, the system and the user talk about different objects and relations. Each part of the dialogue thus may be mapped to an object or a relation in the conversational
world or may be mapped 
to the world itself (grey). In the example, the first part (blue) is about \textit{Object 1} of type \textit{hotel}. When the focus shifts towards \textit{Object 2} of type \textit{restaurant} (green) at U3, the user also addresses the relation (red) \textit{in the same area} between \textit{Object 1} and \textit{Object 2}.

Addressing a relation in this way could still be captured by the semantic interpretation of the user input as the information \emph{area=centre} may be derived from the context. However, if the user said \emph{I need a hotel and a restaurant in Cambridge in the same area} right in the beginning of the dialogue (U1), no context information would be available. To capture these dialogue structures, the dialogue model and the corresponding dialogue state must be able to represent them adequately.

The proposed CEDM achieves this by modelling state information about conversational entities instead of domains. More precisely, it models separate states about the objects (e.g., the hotel or restaurant) and the relations. Previous work on dialogue modelling already incorporated the idea of objects or entities to be the principal component of the dialogue state~\cite{grosz1977representation,bilange1991approach,montoro2004plug,xu2010dialogue,heinroth2013}. However, these dialogue models are not based on statistical dialogue processing where a probability distribution over all dialogue states needs to be modelled and maintained. This additional complexity, though, cannot be incorporated in a straight-forward way into the proposed models. In contrast, the CEDM offers a comprehensive and consistent way of modelling these probabilities by defining and maintaining entity-based states. Work on statistical dialogue state modelling~\cite{young2010his,lee-stent:2016:SIGDIAL,schulz2017frame} also contain a variant of objects but is still based on the MDDM thus not offering any mechanism to model multiple entities or relations between objects. \citet{ramachandran2015belief} proposed a belief tracking approach using relational trees. However, they only consider static relations present in the ontology and are not able to handle dynamic relations.




\section{Statistical Spoken Dialogue Systems}
\label{sec:ssds}
\begin{figure}[t]
\begin{center}
  \includegraphics[width=\linewidth]{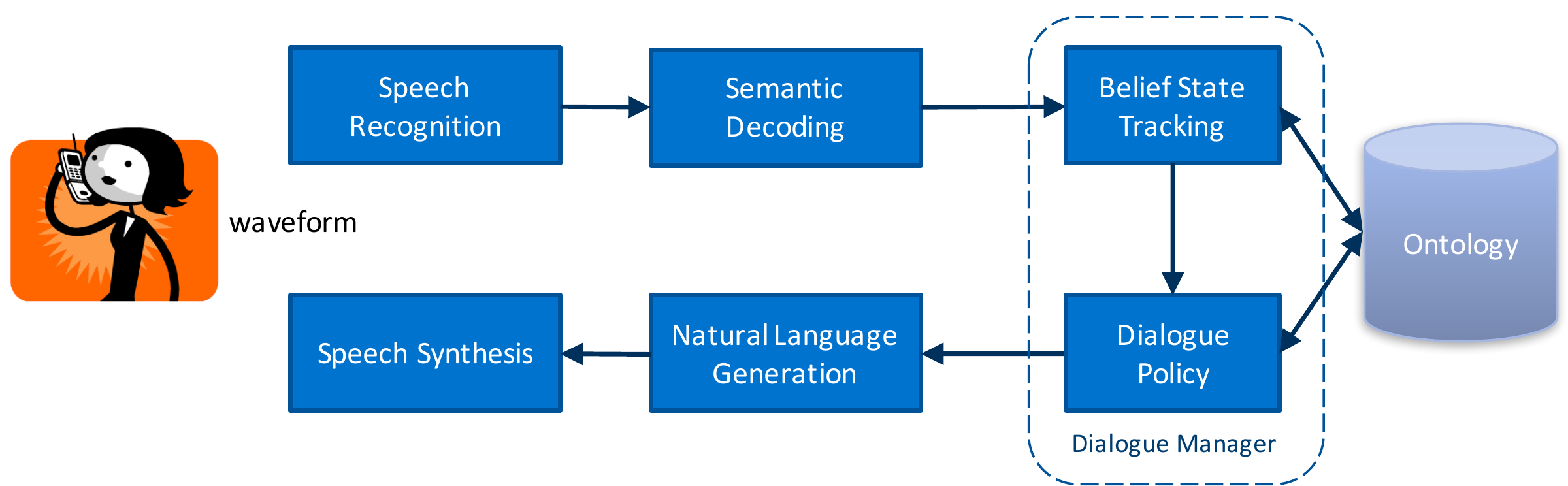}
  \caption{The modular statistical dialogue system architecture. The dialogue manager takes the semantic interpretation as input to track the belief state. The updated state is then used by the dialogue policy to decide on the next system action.}
  \label{fig:ssds_overview}
\end{center}
\end{figure}

Statistical SDS are model-based approaches\footnote{Model-free approaches like end-to-end generative networks~ \cite{serban2016building,li2016} have interesting properties (e.g., they only need text data for training) but they still seem to be limited in terms of dialogue structure complexity (not linguistic complexity) in cases where content from a structured knowledge base needs to be incorporated. Approaches where incorporating this information is learned along with the system responses based on dialogue data~\cite{eric-manning:2017:SIGDIAL} seem hard to scale.} 
and usually assume a modular architecture (see Fig.~\ref{fig:ssds_overview}). The problem of learning the next system action is  framed as a partially-observable Markov decision process (POMDP) that accounts for the uncertainty inherent in spoken communication. This uncertainty is modelled in the belief state $b(s)$ 
representing a probability over all states $s$.

Reinforcement learning (RL) is used in such a sequential decision-making process where the decision-model (the policy $\pi$) is trained based on sample data and a potentially delayed objective signal (the reward $r$)~\cite{sutton1998}. 
The policy selects the next action $a \in A$ based on the current system belief state $b$ to optimise the accumulated future reward $R_t$ at time $t$:
\begin{equation}
\label{eq:acR}
R_t = \sum_{k=0}^{\infty} \gamma^k r_{t+k+1} \; .
\end{equation}
Here, $k$ denotes the number of future steps, $\gamma$ a discount factor and $r_\tau$ the reward at time $\tau$.

The $Q$-function models the expected accumulated future reward $R_t$ when taking action $a$  in belief state $b$ and then following policy $\pi$:
\begin{equation}
\label{eq:qfunction}
Q^{\pi}(b,a) = E_{\pi}[R_t | b_t = b, a_t = a] \; .
\end{equation}

For most real-world problems, finding the exact optimal $Q$-values is not feasible. Instead, RL algorithms have been proposed for dialogue policy learning based on approximating the $Q$-function directly or employing the policy gradient theorem~\cite{williams2006,daubigney2012comprehensive,gasic2014gaussian,williams-asadi-zweig:2017:Long,su-EtAl:2017:SIGDIAL,casanueva2017benchmarking,papangelis2017single}.

Aside from the policy model, the dialogue model plays an important role: it defines the structure and internal links of the dialogue state as well as the system and user acts (i.e., the semantics the system can understand). Thus, the policy model is only able to learn system behaviour based on what is defined by the dialogue model. By defining the dialogue state, the dialogue model further represents an abstraction over the task ontology or knowledge base restricting the view on the information that is relevant so that the system is able to converse\footnote{Using the knowledge base directly to model the (noisy) dialogue state~\cite{pragst2015,meditskosDasiopoulouPragstUltesVrochidisKompatsiarisWanner2016} usually results in high access times.}
.  Most current dialogue models are built around \textit{domains} which encapsulate all relevant information as a section of the dialogue state that belongs to a given topic, e.g., finding a \textit{restaurant} or \textit{hotel}. However, the resulting flat state that is widely used~\cite[e.g.]{williams2005b,young2010his,thomson2010,lee-stent:2016:SIGDIAL,schulz2017frame} 
is not intuitive to model complex dialogue structures like relations. 

To overcome this limitation, we propose the conversational entity dialogue model which will be described in detail in the following section.








\section{Conversational Entity Dialogue Model}
\label{sec:cedm}
The conversational entity dialogue model (CEDM) is proposed as an alternative way of statistical dialogue modelling having the concept of entities at the core of the model. Entities being objects or relations offer an intuitive way of modelling complex task-oriented dialogues.

\subsection{Objects and Relations}
Objects are entities of a certain object type (e.g., \emph{Restaurant} or \emph{Hotel}) where each type defines a set of attributes (see Fig.~\ref{fig:dialogue}). This type definition matches the contents of the back-end knowledge base and thus the internal representation of real-world objects. This is similar to the definition of domains. In contrast to domains, though, this notion allows the modelling of multiple objects of the same type within a dialogue as well as the modelling of a type hierarchy which may be exploited during policy learning.

\begin{figure}[t]
\centering
\includegraphics[width=0.9\linewidth]{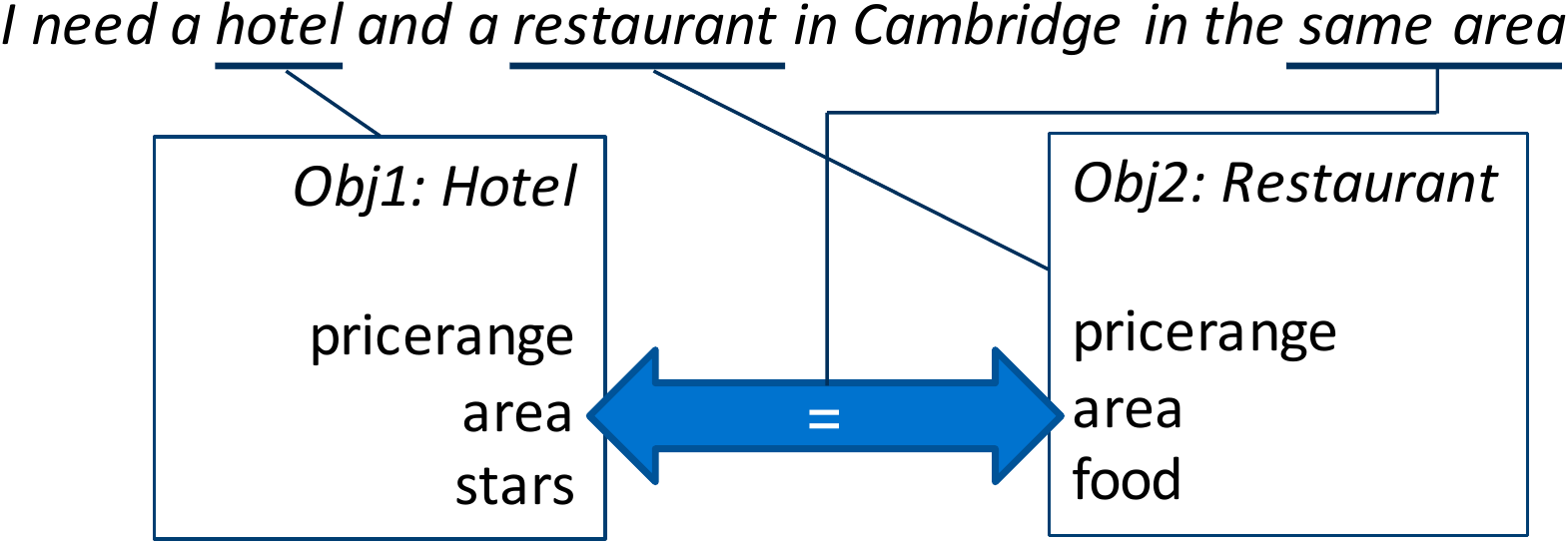}
\caption{Example mapping of a user utterance to two objects and one relation.}
\label{fig:relationexample}
\vspace{-7pt}
\end{figure}

Relations are also entities that connect objects or attributes of objects. An example is shown in Figure~\ref{fig:relationexample}: the two objects \textit{obj1} and \textit{obj2} of types \textit{Hotel} and \textit{Restaurant} respectively are connected through the attribute \textit{area} with the \textit{equals} relation.

Possible relations may directly be derived from the object type definitions, e.g., by allowing only connections for attributes that represent the same concepts like area. Note that these relations are dynamic relations that may be drawn between objects in a conversation. This is different to static relations which are often used in knowledge bases to describe how concepts relate to each other.

\subsection{Conversational Entities in a Conversational World}
A conversational entity is a virtual entity that exists in the context of the current conversation and is either a conversational object or a conversational relation. A conversational object may match a real-world entity but does not need to. In fact, the task of a goal-oriented dialogue is often to find a matching real-world entity based on the information acquired by the system during the dialogue. In the example dialogue (Fig.~\ref{fig:dialogue}), matching entities have already been found for both objects. However, a conversational object exists independently of whether a matching real-world entity has been found yet or even exists. 

\begin{figure}[t]
\centering
\includegraphics[width=\linewidth]{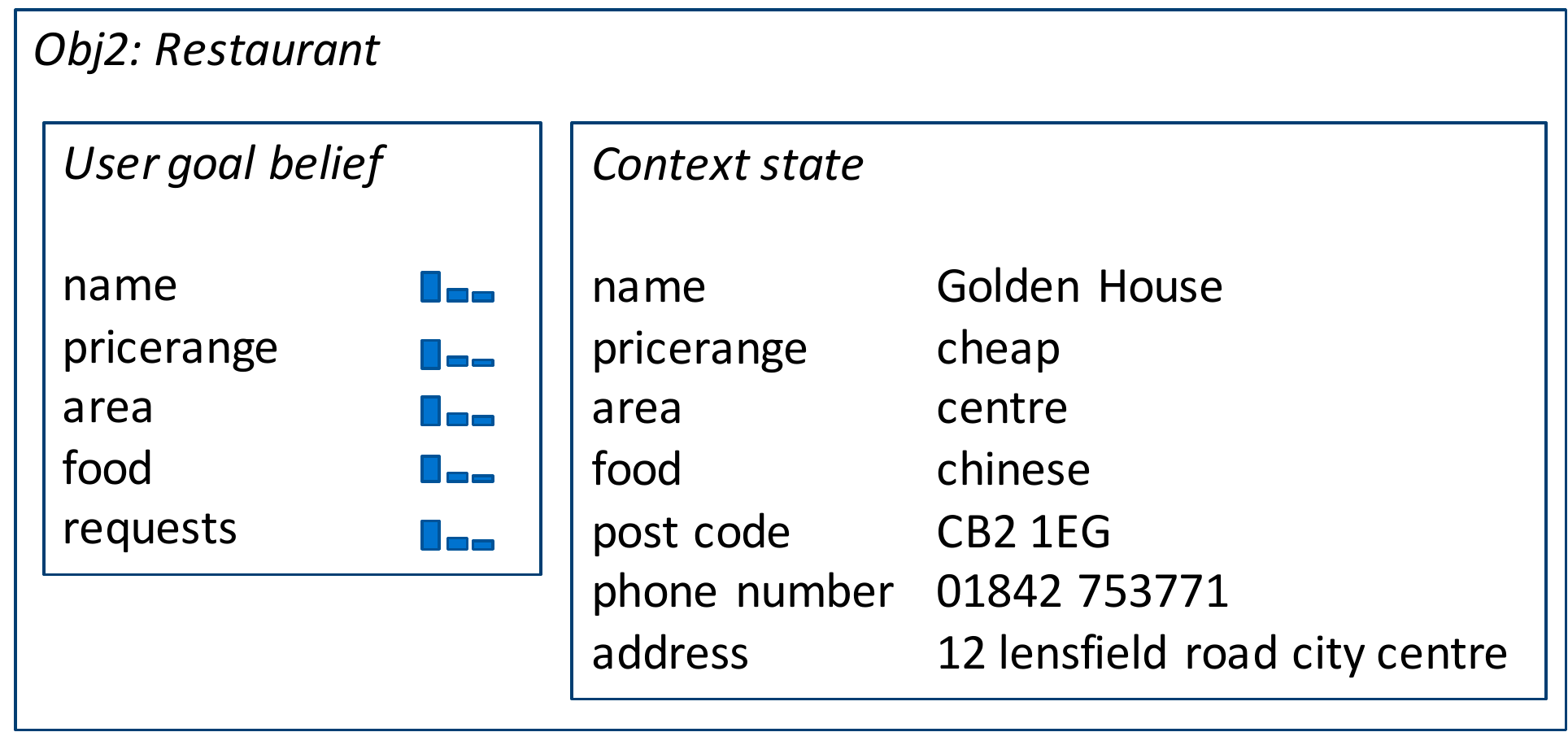}
\caption{Example of a conversational entity representing object \textit{obj2} of type \textit{Restaurant}. The user goal belief models the search constraints the user has provided to the system and the context state represents the most recent real-world match offered by the system.}
\label{fig:ce_obj}
\vspace{-7pt}
\end{figure}

Derived from the object type definition, a conversational object comprises an internal state that consists of the user goal belief $s_u$ and the context state $s_c$ as shown in the example in Figure~\ref{fig:ce_obj}. There, $s_u$ is depicted using marginal probabilities for each slot (which is common in recent work on statistical SDS). While the user goal belief models the system's belief of what the user wants based on the user input, the context state models information that the system has shared with the user. In the example of Figure~\ref{fig:ce_obj}, the system has already offered a matching real-world object based on the user goal belief of the conversational object. If no offer has been made yet, the context state is empty.

The context state plays an important role as addressed relations usually refer to the object offered by the system instead of search constraints represented by the user goal belief. The context state further allows to relate to attributes that have not been mentioned in the dialogue.

\begin{figure}[t]
\centering
\includegraphics[width=\linewidth]{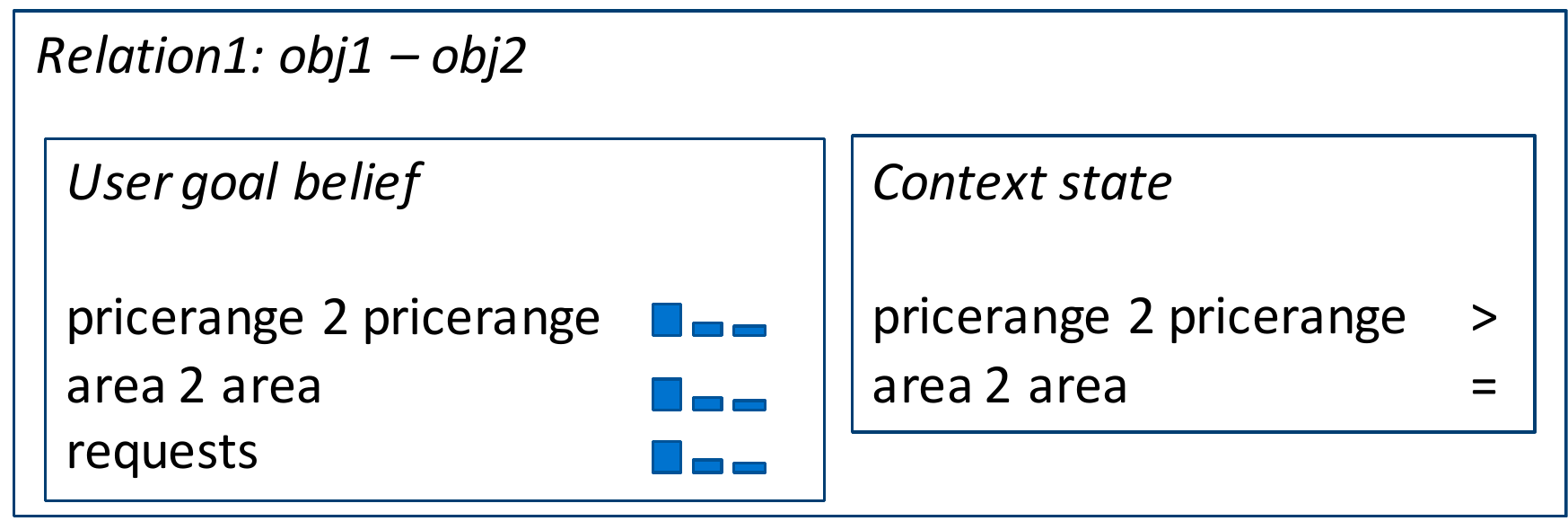}
\caption{Example of the conversational entity \textit{Relation1} between \textit{obj1} and \textit{obj2}. The user goal belief models the search constraints the user has provided to the system and the context state represents the relations based on the most recent real-world matches for both objects offered by the system.}
\label{fig:ce_rel}
\vspace{-8pt}
\end{figure}

One key aspect of the CEDM is that relations are also modelled as a conversational entity. Thus, these conversational relations also define a user goal belief and a context state as shown in Figure~\ref{fig:ce_rel}. The attributes of the relation are created out of the attributes of the objects they connect. In the given example, the attributes \textit{area} and \textit{pricerange} of the two objects are connected resulting in the relation attributes \textit{area2area} and \textit{pricerange2pricerange}. The values of these attributes are the actual relations, e.g., \textit{equals} or \textit{greater/less than}. Similar to the slot belief of conversational objects, each attribute is modelled with a marginal probability over all possible relations.

Assigning part of the belief state to the relations enables the system to specifically react to these relations and even to address them in a system utterance. Furthermore, if the context state of one of the related objects changes (e.g., because the user changed their mind), the relation may still persist.

Each conversational entity resides within a conversational world $w$ (see Fig.~\ref{fig:dialogue}) that defines the number of objects and the type of each object (relations may be derived from this) as well as general state information. This world may either be predefined or needs to be derived from the user input. In the latter case, the user input is usually noisy and an uncertainty needs to be modelled within the dialogue state. As this work focuses on relation modelling, a predefined conversational world is used leaving the uncertainty modelling of conversational worlds for future work.



\subsection{Belief Tracking and Focus of Attention}
The task of belief tracking is to update the probability distribution $b'(s)$ over the states $s$ based on the system action $a$, the observation $o$ of the user input and the previous probability distribution $b$:
\begin{equation}
b'(s) = P(s|o,a,b) \; .
\end{equation}
With the additional complexity of the CEDM having an unknown number of entities in a conversational world, we propose to decompose the state $s$ in the spirit of work by \citet{williams2005b}. The belief update for each entity $e$ is then defined as
\begin{equation}
\label{eq:entitybeliefupdate}
b'_{e}(u,s_u,s_c,h_e) = P(u,s_u,s_c,h_e|a,o,b_e) \; ,
\end{equation}
where $s_u$ is the user goal state of entity $e$, $s_c$ the context state of $e$, $h_e$ the dialogue history of $e$ and $u$ the last user action\footnote{In case of an unknown number of entities represented by a probability over worlds, the probability in Equation~\ref{eq:entitybeliefupdate} needs to be extended to depend on the conversational world and needs to be multiplied by a probability over all worlds.}.

The belief update for the world belief $b_w$ is 
\begin{equation}
b'_w(u,s_w,h_w) = P(u,s_w,h_w|a,o,b_w) \; ,
\end{equation}
where $s_w$ is the world state of world $w$, $h_w$ the dialogue history and $u$ the last user action.

This multi-part belief allows hierarchical dialogue processing on the world level and the entity level as depicted in Figure~\ref{fig:universelevels}. Each level produces its own belief and based on that, the system is able to act on each level. On the world level, the system might produce general dialogue behaviour like greetings or engage in a dialogue to adequately identify the entity which is addressed by the user input. On the entity level, the system talks to the user to acquire information about the concrete entity the user is talking about, e.g., to find a matching entity in the knowledge base. 

In addition to belief tracking, we would like to introduce another concept called focus of attention. Based on work by \citet{grosz1978}, we define the current focus of attention $\mathcal{F}$ for each conversational world as a subset of conversational entities in this world $\mathcal{F} \subseteq W$. Hence, the task of focus tracking is to find the new set of conversational entities which is in the current focus of attention based on the user input and the updated belief state. Even though the concept of focus is not mandatory, it may be helpful when framing the reinforcement learning problem as it allows to limit the size of the input to the reinforcement learning algorithm as well as the number of actions available to the learning algorithm at a given time. Using $\mathcal{F}$ may also prevent the system from acting in parts of the belief state that are completely irrelevant to the current part of the conversation.





\subsection{The Conversational Entity vs.\ the Multi-Domain Dialogue Model}
\label{ssec:cedmvsmddm}

\begin{figure}[t]
\small
\centering
\begin{tabular}{ccc}
\toprule
world level & world general behaviour & $b_w$ \\
\midrule
entity level & entity specific behaviour & $b_e$ \\
\bottomrule
\end{tabular}
\caption{The layered model of the CEDM with the respective components of the belief state.}
\label{fig:universelevels}
\end{figure}

The functionality and the modelling possibilities of the proposed CEDM go beyond (and thus include) the possibilities of the multi-domain dialogue model (MDDM). To demonstrate this, we will outline how a dialogue using the MDDM may be modelled using the CEDM. The core concept \textit{domain} of the MDDM may be mapped to one conversational object of a specific type where the slots of the domain are the attributes of the type. Since the number of domains is predefined, there is only one conversational world with a set number of conversational objects. Relations may not be modelled using the MDDM. Belief update is reduced to finding the right entity for the user input and updating its state. In the CEDM, the semantic decoding of user input includes the entity (or entity type) it refers to, which is similar to the topic tracker of the MDDM where the topic tracker also defines the domain the system acts in. Hence, the focus of attention will always contain only the entity that has been addressed by the user. By that, a policy for each conversational object (and thus object type) may be trained which is the same as the domain policies of the MDDM.

\section{Relation Modelling Evaluation}
\label{sec:evaluation}

To demonstrate the capabilities and benefits of the conversational entity dialogue model (CEDM), the aspect of relation modelling has been selected as it is a core concept of the CEDM. For this, we built upon the mapping to the multi-domain dialogue model (MDDM) as described in Section~\ref{ssec:cedmvsmddm} and extend it with conversational relations. After a brief description of the model implementation, the experiments and their results are presented using two conversational objects of different types. Note that only the \textit{equals} relation is considered here due to limitations of the marginal belief state model.

\subsection{Model Implementation}
To implement all relevant aspects of the CEDM, the publicly available open-source statistical dialogue system toolkit PyDial~\cite{ultes2017pydial} is used which originally follows the MDDM. 

The main challenge for policy implementation is to integrate both the state of the object in $\mathcal{F}$ as well as the states of all corresponding relations into the dialogue decision. To achieve this, a hierarchical policy model based on feudal reinforcement learning~\cite{dayan1993feudal} has been implemented following the approach of~\citet{casanueva2018feudal}. For each object type, a master policy decides whether the next system action addresses a conversational relation or the conversational object. A respective sub-policy is then invoked in a second step where each object type and each relation type are modelled by an individual policy. Thus, the model decomposes the action selection problem to take account for the specificities of the object policy and relation policies respectively and is able to handle a variable number of relations and a large state space. During training, all policies (master and sub-policies) receive the same reward signal.  

Aside from the feudal RL architecture which seems to be intuitive for the proposed CEDM, 
the main problem is the handling of back-end data-base access. In the MDDM, each domain models all information which is necessary to do the data-base lookup. However, this is not possible in the CEDM as information from different conversational objects and relations need to be taken into account. One way of doing this is to apply a rule-based merging of the state of the conversational object in $\mathcal{F}$ with the states of all other conversational objects that are related through a conversational relation to form the focus state $\hat{b}$:
\begin{equation}
\hat{b}_{s}(v) = \frac{\sum_{i} w_i b_s^i(v)}{\sum_{i} w_i} \;  ,
\end{equation}
where $s$ is the slot, $v$ is the value, and $b^i$ the belief of the $i$-th conversational entity involved in the merging process. $w_i = 1-b^i_s(\emptyset)$ is the weight of the $i$-th conversational entity where $b^{i}_s(\emptyset)$ represents the probability where no information about slot $s$ has yet been shared with the system. $b^i$ either refers to the belief $b^o$ of the conversational object $o$ in $\mathcal{F}$ or to an already weighted belief $\tilde{b}^{o'}$ originating from the conversational relation $rel^{oo'}$ connecting conversational object $o$ with $o'$:
\begin{equation*}
\tilde{b}_s^{o'}(v) =
\begin{cases}
    rel_s^{oo'}(=) \cdot b^{o'}_s(v),& v \neq \emptyset\\
    rel_s^{oo'}(=) \cdot b^{o'}_s(v) + rel_s^{oo'}(\emptyset),&              v = \emptyset
\end{cases}
\end{equation*}
where  $b^{o'}$ is the belief of object $o'$. The relation probability $rel$ is $0$ if the slot $s$ has no matching slot in $o'$. Please note that for $b^{o'}_s(v)$, even though we refer to the belief, the context state of $o'$ is used instead if not empty. The focus state is used as input to the master policy as well as the sub-policy of the conversational object.

As an example, consider $b^o_s = [\emptyset: 0.3, v_1: 0.7, v_2: 0.0]$, $b^{o'}_s = [\emptyset: 0.2, v_1: 0.0, v_2: 0.8]$, and $rel_s^{oo'} = [\emptyset: 0.1, =: 0.9]$. This results in $\tilde{b}_s^{o'} = [\emptyset: 0.28, v_1: 0.0, v_2: 0.78]$ and $\hat{b}_{s} = [\emptyset: 0.29, v_1: 0.35,v_2: 0.36]$. This example also shows that conflicts which may exists between the state of the conversational object and the state defined by the relation are visible at this level. To help the policy to learn in this situation, an additional conflict bit is added to the focus belief state as input to the master policy.

The source code of the CEDM implementation is available at \url{http://pydial.org/cedm}.

\subsection{Experimental Setup}
To evaluate the relation modelling capabilities of the CEDM, the task of finding a hotel and a restaurant in Cambridge has been selected (corresponding to the \textit{CamRestaurants} and \textit{CamHotels} domains of PyDial). The corresponding conversational world consists of two conversational objects of types \textit{hotel} and \textit{restaurant} and one conversational relation. Based on the object type definitions, the conversational relation connects the slots \textit{area} and \textit{pricerange} of both objects. 
Using a simulated environment, the goals of the simulated user were generated so that at least one of these two slots is related (i.e., contains the same value).

To test the influence of the user addressing the relation instead of the correct value (e.g., "restaurant in the same area as the hotel" vs.\ "restaurant in the centre"), we have extended the simulated agenda-based user~\cite{schatzmann2009} with a probability $r$ of the user addressing the relation instead of the value. The higher $r$, the more often the user addresses the relation. The user simulator is equipped with an additional error model to simulate the semantic error rate (SER) caused in a real system by the noisy speech channel. 

For belief tracking, an extended version of the focus tracker~\cite{henderson2014second}---an effective rule-based tracker---was used for the conversational entities and the conversational world that also discounts probabilities if the respective value has been rejected by the user. 
As a simulated interaction is on the semantic level, no semantic decoder for the relations is necessary. For training and evaluation of the proposed framework, both the master policy and all sub-policies are modelled with the GP-SARSA algorithm~\cite{gasic2014gaussian}. 
This is a value-based method that uses a Gaussian process to approximate the state-value function (Eq.~\ref{eq:qfunction}). As it takes into account the uncertainty of the estimate, it is sample-efficient.

To compare the dialogue performance of the CEDM with the MDDM baseline, two experiments have been conducted. All dialogues follow the same structure: the user and the system first talk about one conversational object before moving on to the second object. As the user only addresses a relation to an object that has previously been part of the dialogue, relations are only relevant when talking about the second object. However, there are times where a relation has been addressed by the user before the goal of the first object changed which resulted in the addressed relation being wrong. This could only be resolved by the system by addressing the relation itself.

\vspace{\experimentsgap}
\noindent \textbf{Experiment 1 \;} In the first experiment, the influence of $r$ on the dialogue performance is investigated in a controlled environment. Having a fixed order, only the feudal policy of the second object (where relations may occur), the \textit{restaurant}, is learned. To avoid interfering effects of jointly learning both policies at the same time, the first object \textit{hotel} uses a handcrafted policy.

\vspace{\experimentsgap}
\noindent \textbf{Experiment 2 \;} The second experiment focusses on the joint learning effects. Thus, the order of objects is alternated, all objects use the feudal policy model and are trained simultaneously.

\subsection{Results}

\begin{table*}[t]
  \centering
  \footnotesize
  \caption{\footnotesize Reward and success rate of both experiments for different relation probabilities $r$ comparing the proposed CEDM to the MDDM baseline. The measures only show the performance of the second object in the dialogue where the relation is relevant. All results are computed after 4,000/1,000 train/test dialogues and averaged over 5 trials with different random seeds. \textbf{Bold} indicates statistically significant outperformance ($p < .05$), \textit{italic} indicates no statistically significant difference.\hspace{0.8mm}}
\setlength{\tabcolsep}{3pt}
\begin{tabular}{ccccccccccccccccc}
\toprule
          
          & \multicolumn{8}{c}{Experiment 1} & \multicolumn{8}{c}{Experiment 2} \\
          \cmidrule(l{.25em}r{.25em}){2-9} \cmidrule(l{.25em}r{.25em}){10-17}
          & \multicolumn{4}{c}{Restaurant - Env. 1} & \multicolumn{4}{c}{Restaurant - Env. 3} & \multicolumn{4}{c}{Restaurant  - Env. 3} & \multicolumn{4}{c}{Hotel - Env. 3} \\
          \cmidrule(l{.25em}r{.25em}){2-5} \cmidrule(l{.25em}r{.25em}){6-9} \cmidrule(l{.25em}r{.25em}){10-13} \cmidrule(l{.25em}r{.25em}){14-17}
          
          & \multicolumn{2}{c}{CEDM} & \multicolumn{2}{c}{base} & \multicolumn{2}{c}{CEDM} & \multicolumn{2}{c}{base} & \multicolumn{2}{c}{CEDM} & \multicolumn{2}{c}{base} & \multicolumn{2}{c}{CEDM} & \multicolumn{2}{c}{base} \\
\cmidrule(l{.25em}r{.25em}){2-3} \cmidrule(l{.25em}r{.25em}){4-5} \cmidrule(l{.25em}r{.25em}){6-7} \cmidrule(l{.25em}r{.25em}){8-9} \cmidrule(l{.25em}r{.25em}){10-11} \cmidrule(l{.25em}r{.25em}){12-13} \cmidrule(l{.25em}r{.25em}){14-15} \cmidrule(l{.25em}r{.25em}){16-17}
    $r$   & Rew.  & Suc.  & Rew.  & Suc.  & Rew.  & Suc.  & Rew.  & Suc.  & Rew.  & Suc.  & Rew.  & Suc.  & Rew.  & Suc.  & Rew.  & Suc.  \\
    \midrule
    0.0   & \textbf{23.3} & 99.3\% & 23.2  & \textbf{99.6\%} & 20.4  & 94.3\% & \textbf{20.8} & \textbf{96.6\%} & 20.1  & 95.0\% & \textbf{20.7} & \textbf{96.1\%} & \textit{16.5} & \textbf{86.7\%} & \textit{16.6} & 85.8\% \\
    0.1   & 23.1  & \textbf{99.5\%} & \textbf{23.2} & 99.1\% & 20.5  & 94.7\% & \textbf{21.1} & \textbf{96.5\%} & \textit{20.3} & \textit{94.4\%} & \textit{20.4} & \textit{94.4\%} & 16.5  & 86.4\% & \textbf{17.5} & \textbf{89.0\%} \\
    0.3   & \textbf{23.2} & \textbf{99.5\%} & 23.1  & 99.0\% & 20.2  & 93.6\% & \textbf{21.0} & \textbf{95.8\%} & 19.7  & 93.6\% & \textbf{20.4} & \textbf{95.0\%} & \textit{16.2} & 85.5\% & \textit{16.5} & \textbf{87.1\%} \\
    0.5   & \textbf{22.8} & \textbf{99.6\%} & 21.9  & 96.2\% & \textbf{19.8} & \textbf{92.8\%} & 18.7  & 89.7\% & \textbf{19.7} & \textbf{92.5\%} & 19.3  & 92.0\% & 14.6  & 80.8\% & 15.2  & \textbf{82.4\%} \\
    0.7   & \textbf{22.6} & \textbf{99.2\%} & 17.4  & 82.3\% & \textbf{19.9} & \textbf{92.9\%} & 17.7  & 86.8\% & \textbf{19.2} & \textbf{91.6\%} & 17.9  & 87.9\% & \textbf{16.7} & \textbf{86.9\%} & 12.7  & 75.7\% \\
    0.9   & \textbf{22.5} & \textbf{99.4\%} & 5.3   & 41.6\% & \textbf{19.3} & \textbf{91.2\%} & 15.0  & 79.8\% & \textbf{18.2} & \textbf{89.5\%} & 14.2  & 78.2\% & \textbf{9.8} & \textbf{64.3\%} & 8.1   & 61.5\% \\
    1.0   & \textbf{21.6} & \textbf{99.5\%} & -3.6  & 11.7\% & \textbf{18.9} & \textbf{90.2\%} & 13.9  & 76.8\% & \textbf{17.9} & \textbf{88.3\%} & 10.9  & 67.5\% & \textbf{13.8} & \textbf{79.4\%} & 7.0   & 58.2\% \\
    \bottomrule
    \end{tabular}%
    
  \label{tab:results0}%
\end{table*}%

\begin{figure*}[t]
  \centering
  \begin{minipage}[b]{0.495\linewidth}
    \centering
    \includegraphics[width=\linewidth,height=5cm]{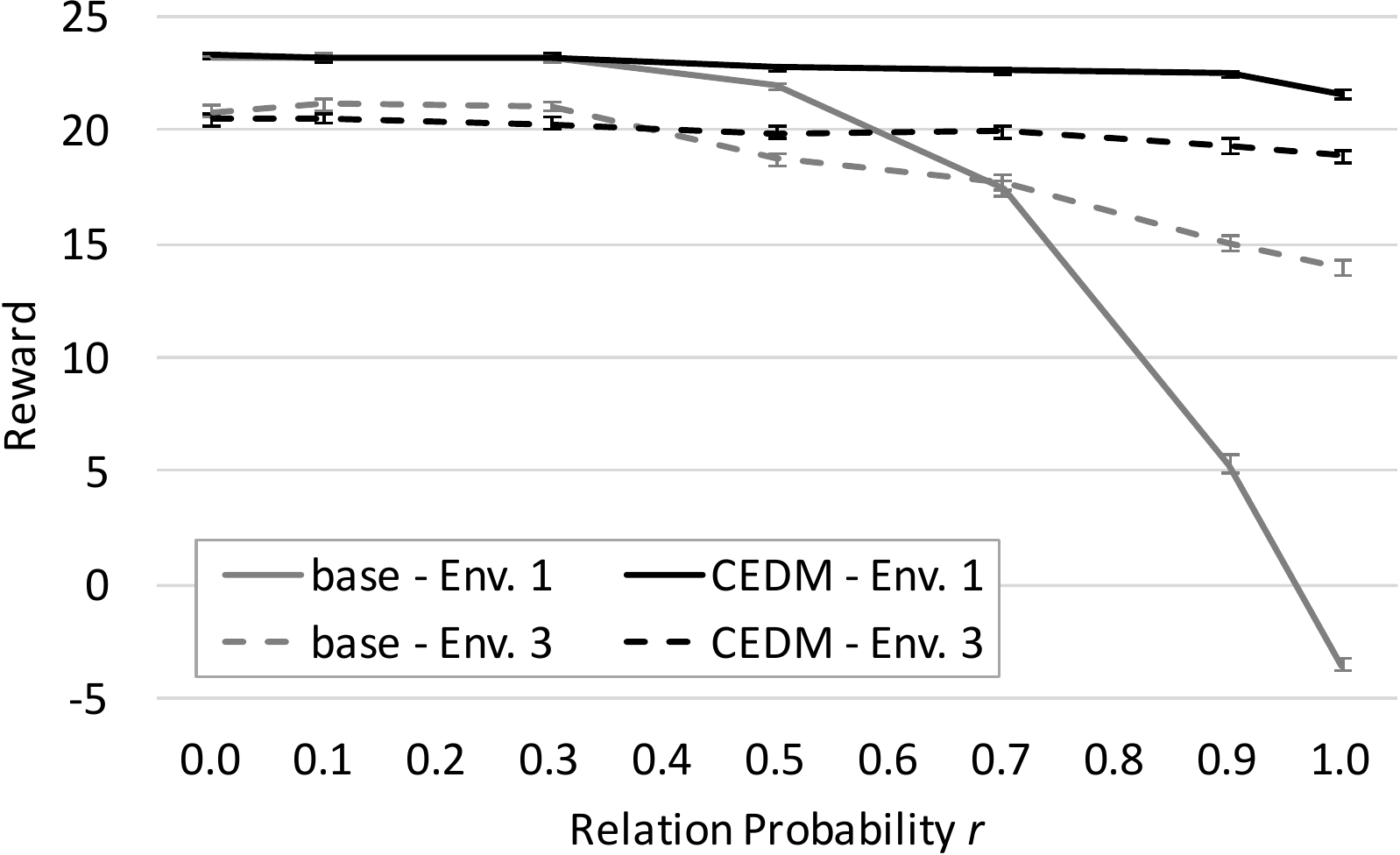}
  \end{minipage}%
  \hfill
  \begin{minipage}[b]{0.495\linewidth}
    \centering
    \includegraphics[width=\linewidth,height=5cm]{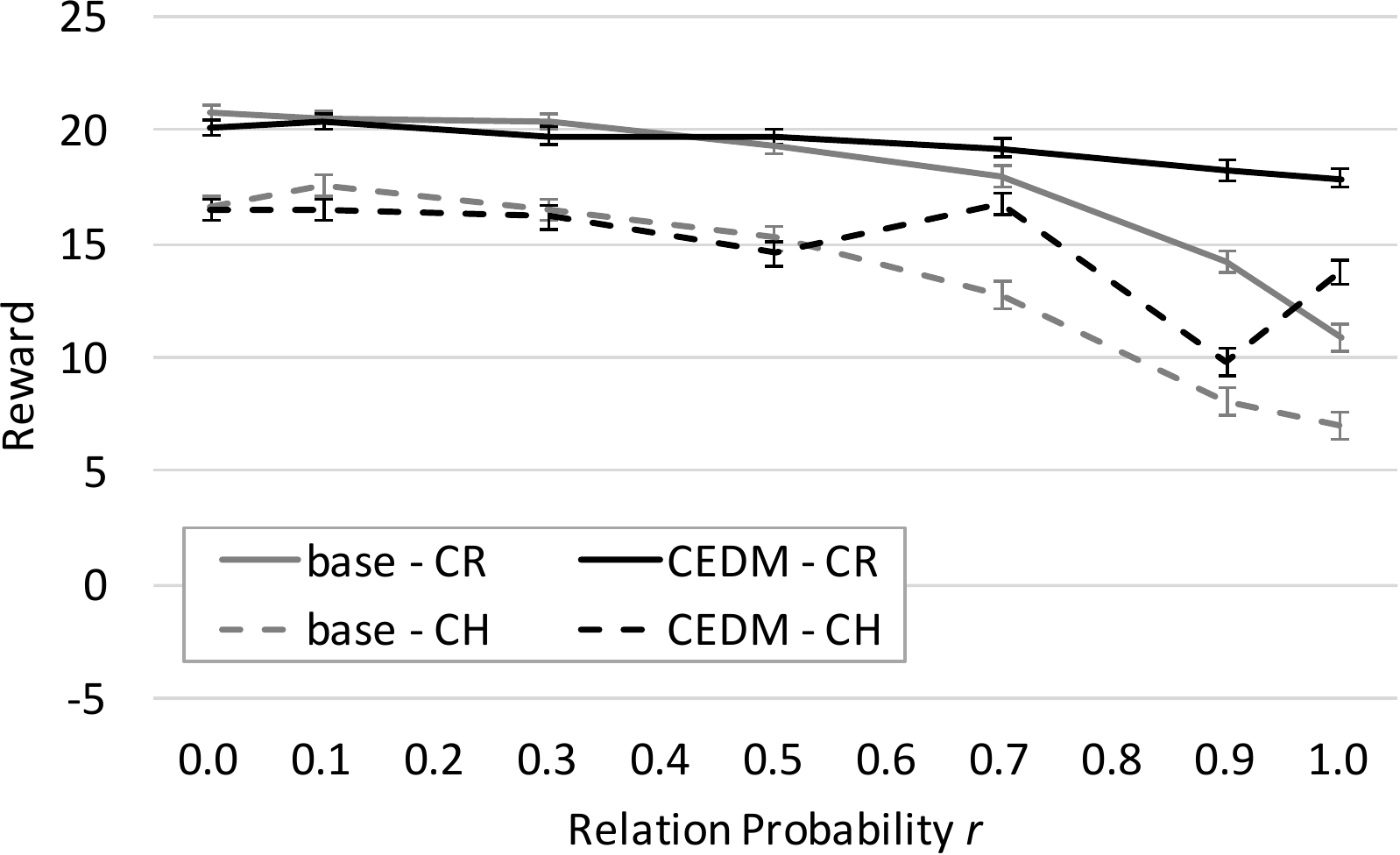}
  \end{minipage}%
\vspace{-6pt}
  \caption{\footnotesize Reward and confidence interval of Experiment~1 (left) and Experiment~2 (right) for different relation probabilities $r$ comparing the proposed CEDM to the MDDM baseline. The measures only show the performance of the second object in the dialogue where the relation is relevant. All results are computed after 4,000/1,000 train/test dialogues and averaged over 5 trials with different random seeds.\hspace{0.5mm}}
  \label{fig:result}
  \vspace{-3pt}
\end{figure*}

The experiments have been conducted based on the PyDial simulation environments Env.~1 and Env.~3 specified by \citet{casanueva2017benchmarking} where Env.~1 operates on a clean communication channel with an SER of 0\% and Env.~3 simulates an SER of 15\%. For each experiment, a policy for the respective object types was trained with 4,000 and tested with 1,000 dialogues. The reward was set to +30/+0 for success/failure and -1 for each turn with a maximum of 25 turns per object. The results were averaged over 5 different random seeds. 

\vspace{\experimentsgap}
\noindent \textbf{Experiment 1 \;} As can be seen in Table~\ref{tab:results0} and Figure~\ref{fig:result} on the left, the proposed CEDM with a feudal policy model is easily able to deal with relations addressed by the user for any relation probability $r$ in both environments. Success rate and reward achieve similar results for all $r$. Only for very high $r$, a small reduction in performance is visible. This can be explained with the added complexity of the dialogue itself as well as the system actions that address the relations. A high relation probability for a slot requires the system to address either the relation or the slot value directly. Both actions may have similar or contradicting impact on the dialogue which makes it harder to learn a good policy. In Env.~3, the added noise results in minor fluctuations which may be expected. 

In contrast, the baseline (the MDDM) is not able to handle the user addressing relations adequately for higher $r$: while for low $r$, the policy is able to compensate by requesting the respective information again, the performance drops at around $r = 0.5$. The reason why the performance of the baseline does not drop as much in Env.~3 as it does in Env.~1 is the way the simulated error model of the simulated user operates. By producing a 3-best-list of user inputs, the chance that the actual correct value is introduced as noise if a relation has originally been uttered is relatively high. As the n-best-list of Env.~1 has the length of one, this does not happen there. 

The performance of the hand-crafted hotel policy was similar for all $r$ in Env.~1 with $rew = 23.4, suc = 99.7\%$ and in Env.~3 with $rew = 20.1, suc = 94.5\%$. 

Analysing the system actions of the dialogues of the CEDM shows that the system learns to address a relation in up to 28\% of all dialogues for $r = 1.0$. 

Example dialogues for Env.~1 are shown in Figures~\ref{dialg:cedm} and \ref{dialg:mddm}.

\vspace{\experimentsgap}
\noindent \textbf{Experiment 2 \;} The results shown in Table~\ref{tab:results0} and Figure~\ref{fig:result} on the right show the performance of the conversational object policies when the respective object was the second one in the dialogue (where relations occur). Still, policies of both objects were trained in all dialogues. The effects of this added noise become visible in the results as they seem to be less stable. Furthermore, the overall performance for the \textit{restaurant} policy drops a bit, but still shows the same characteristics as in Experiment~1. Learning a \textit{hotel} policy results in worse overall performance (which matches the literature) and in cases where a relation is involved.

The performance of the policy of the first object was similar for all $r$ where the restaurant policy achieved $rew = 21.5, suc = 95.4\%$ and the hotel policy $rew = 18.8, suc = 90.2\%$.

Analysing the system actions of the dialogues shows that the CEDM learns to address a relation in up to 24.5\% of all dialogues for $r = 1.0$. 



\section{Conclusion and Future Work}
\label{sec:conclusion}
In this paper, we have presented a novel dialogue model for statistical spoken dialogue systems that is centred around objects and relations (instead of domains) thus offering a new way of modelling statistical dialogue. The two major advantages of the new model are the capability of including multiple objects of the same type and the capability of modelling and addressing relations between the objects. By assigning a part of the belief state not only to each object but to each relation as well, the system is able to address the relations in a system response.

We have demonstrated the importance of the aspect of relation modelling---a core functionality of our proposed model---in simulated experiments showing that by using a hierarchical feudal policy architecture, adequate policies may be learned that lead to successful dialogues in cases where relations are often mentioned by the user. Furthermore, the resulting policies also learned to address the relation itself in the system response.

However, only a small part of the proposed dialogue model has been evaluated in this paper. To explore its full potential, many questions need to be addressed in future work. For creating a suitable semantic decoder that is able to semantically parse linguistic information about relations, an extensive prior work on named entity recognition and dependency parsing already exists and needs to be leveraged and applied to conduct real user experiments. Moreover, relations other than \textit{equals} need to be investigated. Finally, the challenges of identifying all conversational entities in the dialogue and assigning the correct one to each user action as well as finding suitable belief-tracking approaches for the proposed multi-layered architecture along with effective policy models need to be addressed. 



\section*{Acknowledgments}
This research was partly funded by the EPSRC grant EP/M018946/1 \textit{Open Domain Statistical Spoken Dialogue Systems}. 

\bibliography{references}
\bibliographystyle{acl_natbib_new}

\clearpage
\appendix

\onecolumn

\section{Example Dialogues}
\begin{minipage}{\linewidth} 
\small
\begin{tabularx}{\linewidth}{rlL}
1: & Sys & hello() \\
\rowcolor{Hotel} 2: & User & inform(CamHotels\#type="placetostay", CamHotels\#area="west", CamHotels\#pricerange="moderate", CamHotels\#kind="guesthouse") \\
\rowcolor{Hotel} 3: & Sys & inform(CamHotels\#kind="guesthouse", CamHotels\#pricerange="moderate", CamHotels\#name="hobsons house", CamHotels\#area="west") \\
\rowcolor{Hotel} 4: & User & reqalts() \\
\rowcolor{Hotel} 5: & Sys & inform(CamHotels\#name="none", CamHotels\#name!="hobsons house", CamHotels\#kind="guesthouse", CamHotels\#pricerange="moderate", CamHotels\#area="west") \\
\rowcolor{Hotel} 6: & User & reqalts(CamHotels\#area="north") \\
\rowcolor{Hotel} 7: & Sys & inform(CamHotels\#kind="guesthouse", CamHotels\#pricerange="moderate", CamHotels\#name="limehouse", CamHotels\#area="north") \\
\rowcolor{Hotel} 8: & User & request(CamHotels\#price) \\
\rowcolor{Hotel} 9: & Sys & inform(CamHotels\#name="limehouse", CamHotels\#price="a cheapest single room is 40 pounds and a cheapest double room is 75 pounds") \\
\rowcolor{Restaurant} 10: & User & inform(CamRestaurants\#type="restaurant", CamRestaurants\#food="british") \\
\rowcolor{Restaurant} 11: & Sys & request(CamRestaurants\#pricerange) \\
\rowcolor{Restaurant} 12: & User & inform(CamRestaurants\#pricerange="moderate") \\
\rowcolor{Restaurant} 13: & Sys & request(CamRestaurants\#area) \\
\rowcolor{Restaurant} 14: & User & inform(CamRestaurants\#area=CamHotels\#area) \\
\rowcolor{Restaurant} 15: & Sys & inform(CamRestaurants\#name="none", CamRestaurants\#food="british", CamRestaurants\#pricerange="moderate", CamRestaurants\#area="north") \\
\rowcolor{Restaurant} 16: & User & negate(CamRestaurants\#area="west") \\
\rowcolor{Restaurant} 17: & Sys & confirm(CamHotels\#area=CamRestaurants\#area) \\
\rowcolor{Restaurant} 18: & User & negate(CamRestaurants\#area="west") \\
\rowcolor{Restaurant} 19: & Sys & inform(CamRestaurants\#food="british", CamRestaurants\#pricerange="moderate", CamRestaurants\#name="saint johns chop house", CamRestaurants\#area="west") \\
20: & User & bye() \\
21: & Sys & bye() \\

\end{tabularx}
\figcaption{An example simulated dialogue using the proposed CEDM (Experiment 1, Env.~1, $r=0.7$) about finding a hotel first (blue) followed by a restaurant (green). The system is able to process user input addressing a relation (Turn 14) as well as to address the relation in a system utterance (Turn 17). The dialogue nicely shows how the system is able to resolve a conflict between the belief of the conversational relation and the belief of the conversational object (i.e., CamRestaurants\#area = CamHotels\#area vs.\  CamRestaurants\#area = "west").}%
\label{dialg:cedm}%
\end{minipage}

\vspace{0.8cm}
\noindent
\begin{minipage}{\linewidth} 
\small

\begin{tabularx}{\linewidth}{rlL}
1: & Sys & hello() \\
\rowcolor{Hotel} 2: & User & inform(CamHotels\#type="placetostay", CamHotels\#kind="hotel", CamHotels\#area="north", CamHotels\#stars="2") \\
\rowcolor{Hotel} 3: & Sys & inform(CamHotels\#kind="hotel", CamHotels\#name="lovell lodge", CamHotels\#stars="2", CamHotels\#area="north") \\
\rowcolor{Hotel} 4: & User & confirm(CamHotels\#kind="hotel") \\
\rowcolor{Hotel} 5: & Sys & inform(CamHotels\#name="lovell lodge", CamHotels\#kind="hotel") \\
\rowcolor{Restaurant} 6: & User & inform(CamRestaurants\#type="restaurant", CamRestaurants\#food="chinese", CamRestaurants\#pricerange="expensive") \\
\rowcolor{Restaurant} 7: & Sys & request(CamRestaurants\#area) \\
\rowcolor{Restaurant} 8: & User & inform(CamRestaurants\#area=CamHotels\#area) \\
\rowcolor{Restaurant} 9: & Sys & request(CamRestaurants\#area) \\
\rowcolor{Restaurant} 10: & User & inform(CamRestaurants\#area=CamHotels\#area) \\
\rowcolor{Restaurant} 11: & Sys & request(CamRestaurants\#area) \\
\rowcolor{Restaurant} 12: & User & inform(CamRestaurants\#area=CamHotels\#area) \\
\rowcolor{Restaurant} 13: & Sys & request(CamRestaurants\#area) \\
\rowcolor{Restaurant} 14: & User & inform(CamRestaurants\#area="north") \\
\rowcolor{Restaurant} 15: & Sys & inform(CamRestaurants\#food="chinese", CamRestaurants\#pricerange="expensive", CamRestaurants\#name="hakka", CamRestaurants\#area="north") \\
16: & User & bye() \\
17: & Sys & bye() \\

\end{tabularx}
\figcaption{An example dialogue using the MDDM (Experiment 1, Env.~1, $r=0.7$) about finding a hotel first (blue) followed by a restaurant (green). As the system is not able to process user input addressing a relation it keeps requesting the same information (Turns 7-12: CamRestaurants\#area) over and over until the user finally utters the actual value of the area (Turn 14: CamRestaurants\#area="north") which then leads to a successful dialogue.}%
\label{dialg:mddm}%
\end{minipage}

\end{document}